\documentclass{article}
\usepackage{ijcai24}

\usepackage{times}
\usepackage{soul}
\usepackage{url}
\usepackage[hidelinks]{hyperref}
\usepackage[utf8]{inputenc}
\usepackage[small]{caption}
\usepackage{graphicx}
\usepackage{amsmath}
\usepackage{amsthm}
\usepackage{booktabs}
\usepackage{algorithm}
\usepackage{algorithmic}
\usepackage[switch]{lineno}
\usepackage{amssymb}
\usepackage{subfigure}
\usepackage{multirow}

\newtheorem{theorem}{Theorem}
\newtheorem{assumption}{Assumption}

\title{PAFedFV: Personalized and Asynchronous Federated Learning for Finger Vein Recognition}

\author{
Hengyu Mu$^1$
\and
Jian Guo$^1$\and
Chong Han$^1$\And
Lijuan Sun$^1$\\
\affiliations
$^1$Nanjing University of Posts and Telecommunications\\
\emails
\{1022041005, guoj, hc\}@njupt.edu.cn, sunlijuan\_nupt@163.com
}

\begin{document}

\maketitle

\begin{abstract}
With the increasing emphasis on user privacy protection, biometric recognition based on federated learning have become the latest research hotspot. However, traditional federated learning methods cannot be directly applied to finger vein recognition, due to heterogeneity of data and open-set verification. Therefore, only a few application cases have been proposed. And these methods still have two drawbacks. (1) Uniform model results in poor performance in some clients, as the finger vein data is highly heterogeneous and non-Independently Identically Distributed (non-IID). (2) On individual client, a large amount of time is underutilized, such as the time to wait for returning model from server. To address those problems, this paper proposes a Personalized and Asynchronous Federated Learning for Finger Vein Recognition (PAFedFV) framework. PAFedFV designs personalized model aggregation method to solve the heterogeneity among non-IID data. Meanwhile, it employs an asynchronized training module for clients to utilize their waiting time. Finally, extensive experiments on six finger vein datasets are conducted. Base on these experiment results, the impact of non-IID finger vein data on performance of federated learning are analyzed, and the superiority of PAFedFV in accuracy and robustness are demonstrated.
\end{abstract}

\section{Introduction}

Finger vein recognition, as an emerging biometric recognition, has received wide attention from researchers worldwide \cite{r:1,r:2,r:3}. Thanks to the development of deep learning in recent years, the performance of finger vein recognition based on convolutional neural networks has been improving \cite{r:4}. Recent research reveals that the training of finger vein recognition models depends strongly on a large number of centralized finger vein data. However, in real life, obtaining a large-scale finger vein dataset is very difficult and costly. The user's finger vein data is often distributed in each local client. It is difficult for these clients to individually train an effective finger vein recognition model due to lack of data. Moreover, with the increasing emphasis on privacy protection, clients also find it is challenging to expand their datasets by sharing finger vein data. Therefore, how to utilize the data dispersed in clients to train an effective finger vein recognition model has become a research focus in biometrics.

Federated Learning (FL) \cite{c:5}, as a distributed learning framework with privacy-preserving premise, provides a potential solution to the above problem. FedAvg \cite{c:6} is the classical FL method which has gained wide application in Internet of Things (IoT), medicine, autonomous driving, etc. However, existing FL methods cannot be directly applied to finger vein recognition for the following reasons: 1) finger vein recognition is mostly open-set classification tasks in real life. Unlike traditional FL tasks, its data distribution is not the same between the training and the test sets 2) Unlike most other fields where FL is applied, finger vein data among each client is more heterogeneous. Different acquisition devices lead to large discrepancy in image quality, finger location, and light intensity of finger vein images. And these differences make FL for finger vein recognition more difficult. Due to above limitations, FedFV \cite{r:7} is insofar the only FL method for finger vein recognition. It proposed a FL framework, FedFV, for finger vein recognition to address these restrictions. To an extent, FedFV solved the heterogeneity of non-IID finger vein data and performed open-set testing with good results. Nonetheless, it still has some shortcomings. For example, due to its simpler approach to personalizing settings in the local model, it may result in poor model performance for some clients. Similar to the traditional FL setup, it generates a significant amount of waiting time when clients communicate with servers, waiting for servers to process model aggregation, or waiting for other clients to conduct local training.

\begin{figure}[t]
\centering
\includegraphics[width=0.7\columnwidth]{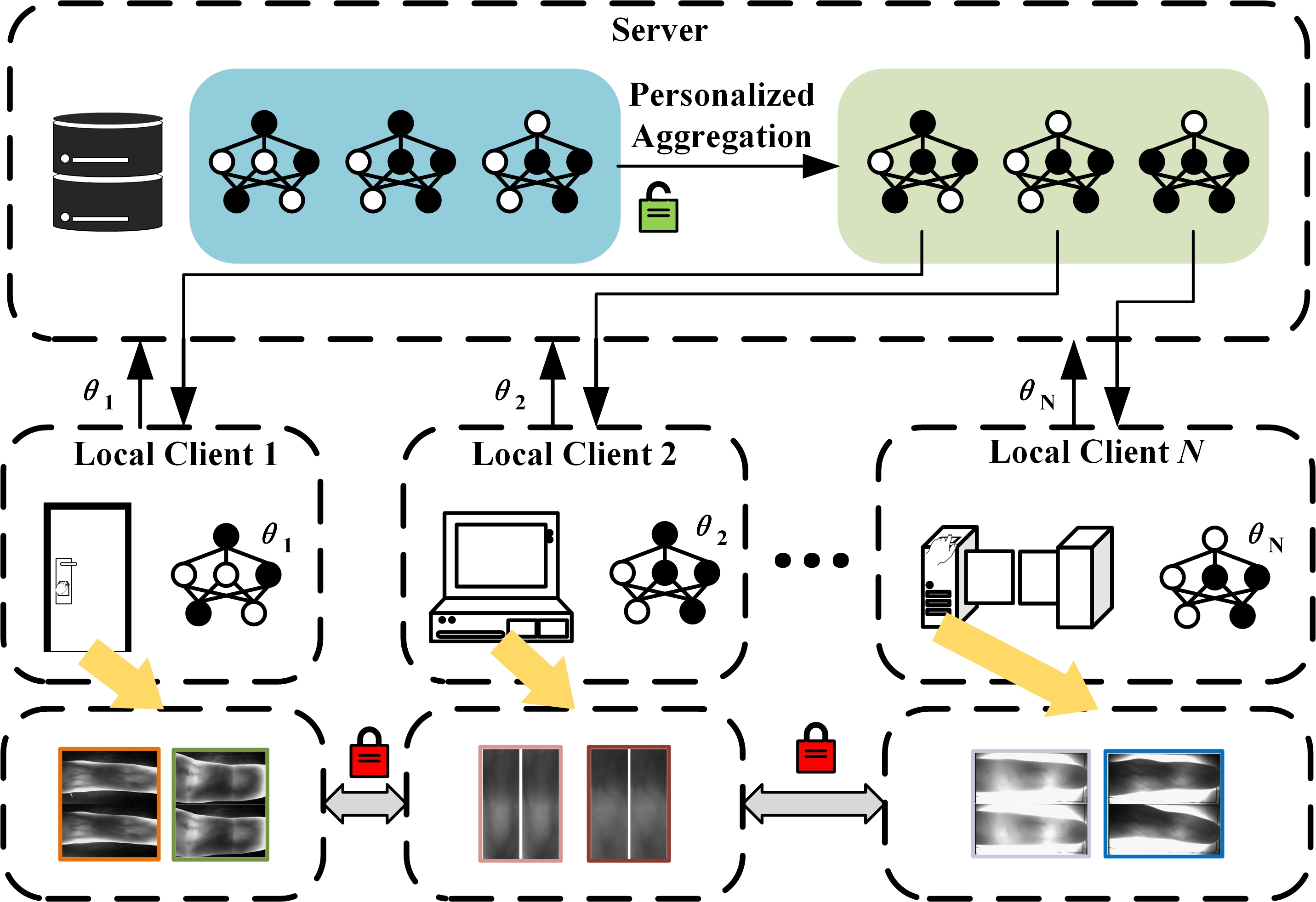} % Reduce the figure size so that it is slightly narrower than the column. Don't use precise values for figure width.This setup will avoid overfull boxes.
\caption{The settings of personalized FL for finger vein recognition. Our goal is to implement a personalized and efficient FL framework to adapt to the highly heterogeneous finger vein data.}
\label{fig_1}
\end{figure}

To address the limitations of the existing work, this paper proposes a Personalized and Asynchronous Federated Learning for Finger Vein Recognition (PAFedFV) framework. The basic setup of PAFedFV is shown in Figure \ref{fig_1}. PAFedFV achieves a more suitable FL for non-IID finger vein data while preserving users’ privacy and security. On the server, it designs a personalized model aggregation method to address the heterogeneity of data among clients. On the clients, it sets up local channel and federated channel to improve the performance of finger vein recognition model. Meanwhile, PAFedFV employs the asynchronous training module to enable the clients utilizing for the waiting time. In addition, this paper conducts extensive experiments on six datasets, FV-USM \cite{r:8}, HKPU-FV \cite{r:9}, NUPT-FV \cite{r:10}, SDUMLA-HMT \cite{c:11}, UTFVP \cite{c:12}, and VERA \cite{c:13}. Through experimental analysis, we identify the patterns of heterogeneous non-IID data that impact finger vein federated learning and demonstrate the superiority of PAFedFV. These findings serve as valuable references for future research and endeavors. In summary, the main contributions of this work are as follows.

\begin{itemize}
    \item We propose a personalized and asynchronous federated learning for finger vein recognition framework, PAFedFV. It enables asynchronous federated learning for non-IID finger vein data while preserving user privacy.
    \item We summarize the patterns of federated learning affected by non-IID finger vein data by analyzing a large number of experiments. These patterns are being revealed for the first time and can serve as references for future research.
    \item Numerous experiments have been conducted on six finger vein datasets. The results demonstrate the superiority of the PAFedFV.
\end{itemize}

\section{Related Work}

\subsubsection {Finger Vein Recognition.} Related researches of finger vein recognition can be mainly categorized into feature-based recognition \cite{r:14,r:15} and deep learning-based recognition \cite{r:16,c:17}. In recent years, deep learn-based finger vein recognition has gradually become crucial thanks to their superior performance. \cite{c:18} proposed SuperPoint-based finger vein recognition (SP-FVR), which utilized the fully convolutional neural network. It solves the problem of poor ability to resist deformation of existing methods, and improves the accuracy and robustness. \cite{r:19} introduced a lightweight finger vein recognition and matching algorithm to reduce computational and time consuming of deep learning models. And \cite{r:20} suggested a low memory overhead and lightweight model for finger vein recognition, which balanced model overhead and performance. These methods have achieved relatively good results, but the majority of them rely on a large amount of finger vein data for model training. However, acquiring a large amount of finger vein data directly is challenging due to privacy protection concerns. This leads to difficulties in the widespread application of finger vein recognition technology. Therefore, it is of practical significance to train an effective model by utilizing scattered finger vein data while protecting user privacy.

\subsubsection {Federated Biometrics.} Federated Learning (FL) is a distributed learning method that coordinates several clients to train an effective model while protecting user privacy. The FegAvg \cite{c:6} is the first algorithm proposed in FL. Since its proposal, FL has drawn attention from researchers in various fields and has achieved significant results \cite{r:25,r:26,r:27}. However, these methods cannot be directly used in finger vein recognition due to the open-set and heterogeneous characteristics of finger vein data.

Recently, the application of federated learning in biometrics has become research focus \cite{c:29,c:31,c:32}. In face recognition, \cite{c:30} proposed the FedFace framework, which combined Partially Federated Momentum (PFM) and Federated Validation (FV). In iris recognition, \cite{c:34} achieved the first federated learning for iris recognition by using iris templates as a communication vehicle. Currently, there are very few studies in finger vein recognition. Feng-Zhao Lian et al \cite{r:7} firstly realized the Federal Learning for Finger Vein Recognition (FedFV) to solve the lack of finger vein data on client. However, FedFV has only simple personalization setup in its model, leading to poor performance on some clients. And like traditional FL, FedFV also generates a large amount of underutilized time for clients. To address above problems, this paper proposes PAFedFV which not only solves the heterogeneity of finger vein data but also makes full use of the waiting time of clients.

\begin{figure*}[t]
\centering
\includegraphics[width=0.58\textwidth]{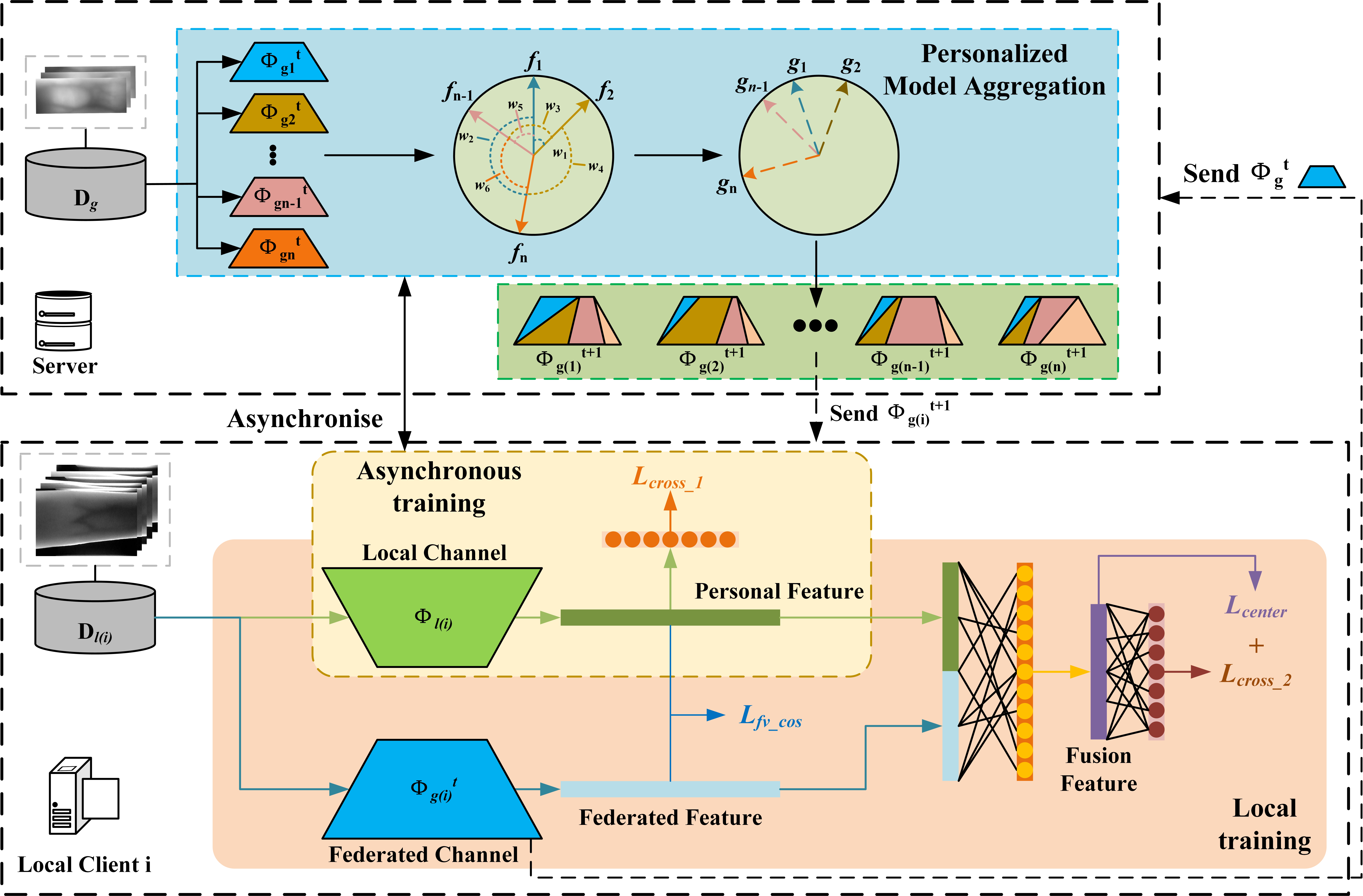}
\caption{The basic framework of PAFedFV. The client trains entire model through local training and uploads the federated channel \emph{$\Phi_{g}$} to server upon completing local training. On server, personalized model aggregation is performed, yielding different global models for each client. Meanwhile, clients engage in asynchronous training for local channel \emph{$\Phi_{l}$} while waiting for the server to return global model.}
\label{fig_2}
\end{figure*}

\section{Proposed Method}
In this work, we propose a novel federated learning framework for finger vein recognition, PAFedFV. The PAFedFV sets up personalized channels within clients' model, thereby improving the recognition performance on clients. On server, a personalized model aggregation method is designed to address the vast heterogeneity among clients' data. In addition, the asynchronized training module is designed to enable each client to fully utilize the waiting time.

\subsection{PAFedFV}
As shown in Figure \ref{fig_2}, the PAFedFV designs a personalized model aggregation method on the server, adapting to non-IID data distribution across clients. And a small finger vein dataset is set up to facilitate the implementation of personalized model aggregation. For each client, we divide the model into two channels: local channel and federated channel, enabling the establishment of a personalized local model. Among them, the local channel remains stored locally and involves in asynchronous training. The federated channel is sent to server for personalized model aggregation after local training. The dual-channel setup enables the local model to strike a balance between the global general features (federated channel) and local private features (local channel). This leads to more comprehensive finger vein representations capable of handling heterogeneous non-IID data. Moreover, to optimize the model of the PAFedFV, four loss functions, \emph{$L_{cross\_1}$}, \emph{$L_{cross\_2}$}, \emph{$L_{center}$} and \emph{$L_{fv\_cos}$} are designed. Among them, \emph{$L_{cross\_1}$} and \emph{$L_{cross\_2}$} are employed to assist the model for training. \emph{$L_{center}$} is used to increase the interclass distance and reduce the interclass distance thereby improving model performance. Lastly, \emph{$L_{fv\_cos}$} serves the purpose of inter-conditioning between federated and local channels, facilitating the extraction of intricate and precise finger vein features. Each module is individually elaborated upon below.

\begin{algorithm}[tb]
\caption{Personalized Model Aggregation}
\label{alg:algorithm_1}
\textbf{Input}: The models \emph{$\Phi_{g(n)}$} from each client, number of clients \emph{N}, A finger vein dataset \emph{I}\\
\textbf{Output}: \emph{N} number of models \emph{${\Phi'}_{g}$}
\begin{algorithmic}[1] %[1] enables line numbers
\STATE Server executes:
\FOR {\emph{n} = 0, 1, …, \emph{N}}
\FOR {\emph{u} = 0, 1, …, \emph{U}(\emph{U}=\emph{N}-\{\emph{n}\})} 
\STATE \emph{${R^{n}}_{u}$} $\gets$ \textit{$\left. {\sum\limits_{t = 1}^{T}\left( \frac{\Phi_{g(n)}\left( I_{t} \right) \bullet \Phi_{g(u)}\left( I_{t} \right)}{\left| \Phi_{g(n)}\left( I_{t} \right) \middle| \times \middle| \Phi_{g(u)}\left( I_{t} \right) \right|} \right.} \right)$}
\ENDFOR
\STATE \emph{$R_{sum}$} $\gets$ \emph{$\sum\limits_{k}^{U}R_{k}^{n}$}
\FOR {\emph{u} = 0, 1, …, \emph{U}(\emph{U}=\emph{N}-\{\emph{n}\})}
\STATE \emph{${\Phi'}_{g(n)}$} = \emph{${\Phi'}_{g(n)}$} + \emph{$\frac{R_{u}^{n} \times \Phi_{g(u)}}{R_{sum}}$}
\ENDFOR
\STATE \emph{${\Phi'}_{g(n)}$} $\gets$ \emph{$\gamma \times {\Phi'}_{g(n)}$} + \emph{$(1-\gamma) \times {\Phi}_{g(n)}$}
\STATE The personalized models \emph{${\Phi'}_{g(n)}$} are sent to \emph{$Client_{n}$}.
\ENDFOR
\end{algorithmic}
\end{algorithm}

\subsection{Personalized Model Aggregation} This section specifies the personalized model aggregation method which instead of the traditional model averaging-based aggregation, which is shown in Algorithm~\ref{alg:algorithm_1}. In traditional FL, model averaging-based aggregation methods exhibit poor performance in finger vein recognition. This occurs because a unified model fails to adjust to distribution variations across diverse finger vein datasets. To address this issue, we have devised a personalized model aggregation method, which is computed as follows. Firstly, the degree of two-by-two correlation between clients' models is calculated. As shown in Equation \ref{eq_1}, the correlation degree between the \emph{n}-th client's and the \emph{u}-th client's models is calculated. 

\begin{equation}
\label{eq_1}
\left. R_{u}^{n} = {\sum\limits_{t = 1}^{T}\left( \frac{\Phi_{g(n)}\left( I_{t} \right) \bullet \Phi_{g(u)}\left( I_{t} \right)}{\left| \middle| \Phi_{g{(n)}}\left( I_{t} \right) \middle| \middle| \times \middle| \middle| \Phi_{g{(u)}}\left( I_{t} \right) \middle| \right|} \right.} \right)~,n \neq u,
\end{equation}
where \emph{$I_{t}$} denotes the \emph{t}-th finger vein image, and \emph{T} denotes the amount of images on the server. \emph{${\Phi}_{g(n)}$} and \emph{${\Phi}_{g(u)}$} denote the model sent by \emph{n}-th client and \emph{u}-th client respectively. $\|.\|$ denotes the inner product computation of vectors. After obtaining all correlation degree values, personalized global models can be aggregated for each client. For example, the \emph{n}-th client's model \emph{${\Phi'}_{g(n)}$} is aggregated by Equation \ref{eq_2}.

\begin{equation}
\label{eq_2}
{\Phi'}_{g(n)} = \gamma \times {\sum\limits_{u}^{U}\left( \frac{R_{u}^{n} \times \Phi_{g(u)}}{\sum\limits_{k}^{U}R_{k}^{n}} \right)} + (1 - \gamma) \times \Phi_{g(n)},
\end{equation}
where \emph{$\gamma$} denotes the threshold set when aggregating. \emph{U} denotes the set of clients except \emph{n}-th client, i.e., $\emph{U}=\emph{N}-\{\emph{n}\}$, \emph{N} denotes the set of all clients. Different global models \emph{${\Phi'}_{g}$} can be aggregated for each client with personalized aggregation.

\begin{figure}[t]
\centering
\includegraphics[width=0.98\columnwidth]{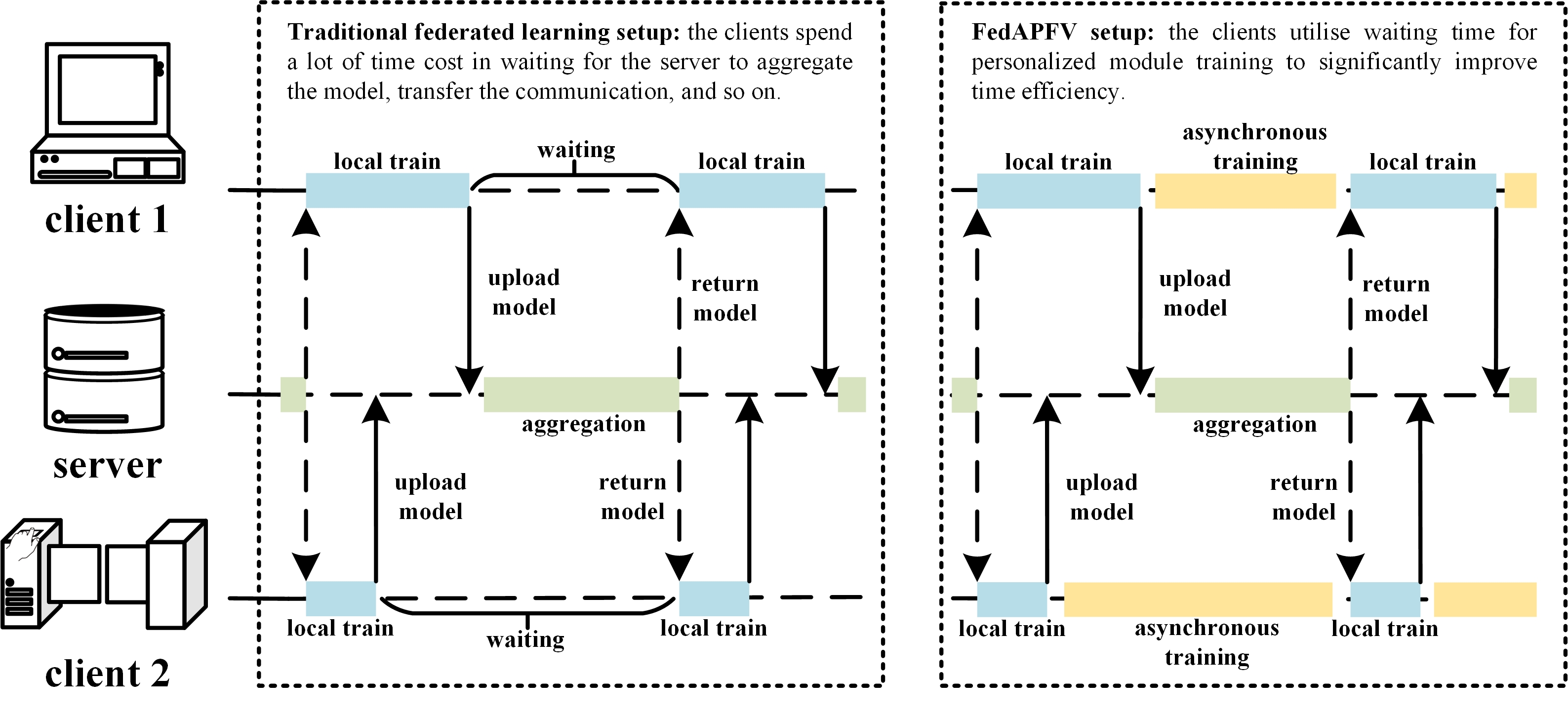}
\caption{The comparison between PAFedFV and traditional federated learning. In traditional federated learning. The clients invest a substantial amount of time while awaiting the return of the global model from the server. In comparison, the PAFedFV employs the asynchronous training module for the client to make full use of the waiting time.}
\label{fig_3}
\end{figure}

\subsection{Asynchronous Training} In traditional federated learning, clients frequently encounter significant waiting periods for the server's response. To address this issue, this paper introduces the asynchronous training module, elaborated upon in detail in this section. As depicted in Figure \ref{fig_3}, in traditional federated learning, clients await the aggregated model from the server before proceeding with their next round of local training. Consequently, this process is synchronous in traditional federated learning, resulting in considerable untapped waiting time for clients. In the PAFedFV framework, instead of idling during the waiting period, clients are allowed to continue training the local channel. Moreover, once the server returns the aggregated model (federated channel), the local and federated channels can be mutually adjusted through \emph{$L_{fv\_cos}$}. And the \emph{$L_{fv\_cos}$} enables the local channel, which continues the training during the waiting time, to guide the optimization of the federated channel. Since the clients can work simultaneously with the server, PAFedFV functions as an asynchronous algorithm. The asynchronous module enables the clients to make full use of the waiting time. And the client defines the \emph{$L_{cross\_1}$} for training the local channel by Equation \ref{eq_3}.

\begin{equation}
\label{eq_3}
L_{cross\_ 1} = - {\sum\limits_{i = 1}^{k}y_{i}} \times log \frac{e^{f_{i}}}{\sum\limits_{j = 1}^{k}e^{f_{j}}},
\end{equation}
where \emph{$y_i$} denotes the one-hot value of the \emph{i}-th image's label. \emph{k} denotes the dimension of the feature \emph{f}, and \emph{$f_i$} denotes the \emph{i}-th value of \emph{f}. And the \emph{f} is calculated by Equation \ref{eq_4}.

\begin{equation}
\label{eq_4}
f = \delta_{1}\left( \Phi_{l}\left( I_{i} \right) \right),
\end{equation}
where \emph{$\delta_{1}$} denotes the linear layer which assists the local channel for training. And we design the \emph{$L_{fv\_cos}$} at local training, which is calculated by Equation \ref{eq_5}. 

\begin{equation}
\label{eq_5}
\left. L_{fv\_ cos} = \middle| \frac{F_{p} \bullet F_{g}}{\left| \middle| F_{p} \middle| \middle| \times \middle| \middle| F_{g} \middle| \right|} - 1 \right|.
\end{equation}

In Equation \ref{eq_5}, \emph{$F_{p}$} and \emph{$F_{g}$} denote the representations which are extracted by \emph{${\Phi}_{l}$} and \emph{${\Phi}_{g}$}, respectively. \emph{${\Phi}_{l}$} and \emph{${\Phi}_{g}$} denote the local and federated channels, respectively. Within it, the local channel could guide the learning of federated channels at local training. And it is able to strike a balance between the local representation with the generic representation for finger vein.

\iffalse 
\begin{algorithm}[tb]
\caption{PAFedFV}
\label{alg:algorithm_2}
\textbf{Input}: The number of clients \emph{n}, the number of local epochs \emph{E}, a finger vein dataset \emph{I}, and learning rate \emph{l}\\
\textbf{Output}: \emph{n} number of models for finger vein recognition
\begin{algorithmic}[1] %[1] enables line numbers
\STATE The client executes:
\WHILE {federated learning is not finished}
\STATE Receive \emph{${\Phi}_{g}$} returned from the server.
\FOR {\emph{e} = 0, 1, …, \emph{E}}
\FOR {\emph{i} = 0, 1, …, \emph{I}}
\STATE The optimization of \emph{${\Phi}_{l}$}, \emph{${\Phi}_{g}$}, \emph{$\delta_{2}$} via local training.
\ENDFOR
\ENDFOR
\STATE The \emph{${\Phi}_{g}$} is sent to server.
\WHILE {server don’t return \emph{${\Phi'}_{g}$}}
\STATE The optimization of \emph{${\Phi}_{l}$} and \emph{$\delta_{1}$} via asynchronous training.
\ENDWHILE
\ENDWHILE
\end{algorithmic}
\end{algorithm}
\fi

\subsection{Learning Pipeline} In this section, the \emph{$L_{total}$} is proposed for optimizing the model at clients. Its specific calculation is defined by Equation \ref{eq_6}.

\begin{equation}
\label{eq_6}
L_{total} = \alpha_{1}L_{fv\_ cos} + \alpha_{2}L_{cross\_ 2} + \alpha_{3}L_{center},
\end{equation}
where \emph{$\alpha_{1}$}, \emph{$\alpha_{2}$}, and \emph{$\alpha_{3}$} denote the thresholds that control the weights of losses, respectively. \emph{$L_{center}$} is proposed by \cite{c:35} for increasing the inter-class distance of features. \emph{$L_{cross\_2}$} represents the cross-entropy loss at local learning, and it is calculated similarly to \emph{$L_{cross\_1}$}. And its feature \emph{f'} is calculated by Equation \ref{eq_7}.

\begin{equation}
\label{eq_7}
f' = \delta_{2}\left( \varphi\left( concat\left\lbrack \Phi_{l}\left( I_{i} \right),\Phi_{g}\left( I_{i} \right) \right\rbrack \right) \right),
\end{equation}
where \emph{$\delta_{2}$}, similar as \emph{$\delta_{1}$}, is a linear layer. And it is used to assist the model in classification and training. \emph{$\varphi$} denotes a deep feature fusion layer. \emph{$I_{i}$} denotes the \emph{i}-th image in the client's finger vein dataset. \emph{$concat\left[\Phi_{g}\left(\emph{$I_{i}$}\right), \Phi_{l}\left(\emph{$I_{i}$}\right)\right]$} denotes the concatenation for the results of two channels.

\section{Convergence Analysis}

In this paper, we explore the federated finger vein recognition task in full device participation, and the proposed PAFedFV framework is an improved implementation base on FedAvg, and our convergence analysis refers to \cite{c:39}.In PAFedFV, we classify functions into personalized local functions \emph{$F'_1$}, · · · , \emph{$F'_N$} and global functions \emph{$F_1$}, · · · , \emph{$F_N$} in terms of whether they perform federated learning. This section is devoted to exploring the convergence of the global functions \emph{$F_1$}, · · · , \emph{$F_N$}. We make the following assumptions on the global functions \emph{$F_1$}, · · · , \emph{$F_N$}. Assumption 1 and 2 are standard because the problem of PAFedFV used softmax classifier. And Assumptions 3 and 4 have been made by the works \cite{r:40,a:41,c:42,c:43}.

\begin{assumption}
\label{as1}
    \emph{$F_1$}, · · · , \emph{$F_N$} are all L-smooth: for all v and w, $F_{k}(v) \leq F_{k}(w) + \left( {v - w} \right)^{T}\nabla F_{k}(w) + \frac{L}{2}\left\| {v - w} \right\|_{2}^{2}$.
\end{assumption}

\begin{assumption}
\label{as2}
    \emph{$F_1$}, · · · , \emph{$F_N$} are all µ-strongly convex: for all v and w, $F_{k}(v) \geq F_{k}(w) + \left( {v - w} \right)^{T}\nabla F_{k}(w) + \frac{\mu}{2}\left\| {v - w} \right\|_{2}^{2}$.
\end{assumption}

\begin{assumption}
\label{as3}
    Let ${\xi^k}_t$ be sampled from the k-th device’s local data uniformly at random. The variance of stochastic gradients in each device is bounded: $\mathbb{E}\left\| {\nabla F_{k}\left( {w_{t}^{k},\xi_{t}^{k}} \right) - \nabla F_{k}\left( w_{t}^{k} \right)} \right\|^{2} \leq \sigma_{k}^{2}$, for k = 1, · · · , N.
\end{assumption}

\begin{assumption}
\label{as4}
    The expected squared norm of stochastic gradients is uniformly bounded, i.e., $\mathbb{E}\left\| {\nabla F_{k}\left( {w_{t}^{k},\xi_{t}^{k}} \right)} \right\|^{2} \leq G^{2}$, for all k = 1, · · · , N and t = 0, · · · , T -1.
\end{assumption}

\begin{theorem}
\label{th1}
    Meanwhile, the aggregation method proposed in this paper satisfies \begin{equation} {\sum\limits_{u}^{U}\frac{\gamma \times R_{u}^{n}}{\sum_{k}^{U}R_{k}^{n}}} + \left( {1 - \gamma} \right) = 1. \end{equation} Then, if $p_{k}$ denotes the weight of k-th client's model, our method satisfies $\sum_{k}p_{k} = 1$.
\end{theorem}

\begin{theorem}
    Let Assumptions \ref{as1} to \ref{as4} hold and combine Theorem \ref{th1}. Then as the proof of \cite{c:39}, the convergence of proposed method satisfies \begin{equation} \mathbb{E}\left\lbrack {F\left( w_{T} \right)} \right\rbrack - F^{*} \leq \frac{2\kappa}{\gamma + T}\left( {\frac{B}{\mu} + 2L\left\| {w_{0} - w^{*}} \right\|^{2}} \right),\end{equation} where \begin{equation} B = {\sum\limits_{k = 1}^{N}{p_{k}^{2}\sigma_{k}^{2} + 6L\left( F^{*} - {\sum_{k = 1}^{N}{p_{k}F_{k}^{*}}} \right)}} + 8(E - 1)^{2}G^{2}. \end{equation}
\end{theorem}

\section{Experiments}
This paper conducts extensive experiments on six datasets, which assesses the efficacy of PAFedFV. This section outlines the experimental setup and presents the results.

\iffalse 
\begin{figure}[t]
\centering
\includegraphics[width=0.7\columnwidth]{fig_4} % Reduce the figure size so that it is slightly narrower than the column. Don't use precise values for figure width.This setup will avoid overfull boxes.
\caption{The sample images of datasets are shown. where (a) is a sample of NUPT-FV (b) is of SDUMLA (c) is of VERA (d) is of USM (e) is of HKPU (f) is of UTFVP.}
\label{fig_4}
\end{figure}
\fi

\begin{figure*}[t]
\centering
\subfigure[]{\label{fig_5a}
\includegraphics[width=0.25\textwidth]{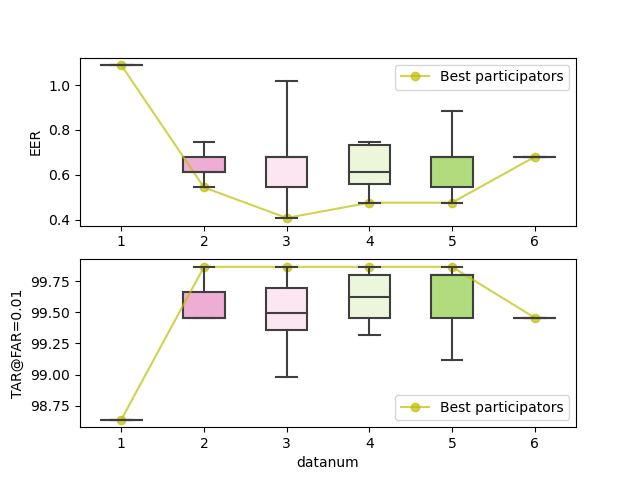}}    
\subfigure[]{\label{fig_5b}     
\includegraphics[width=0.25\textwidth]{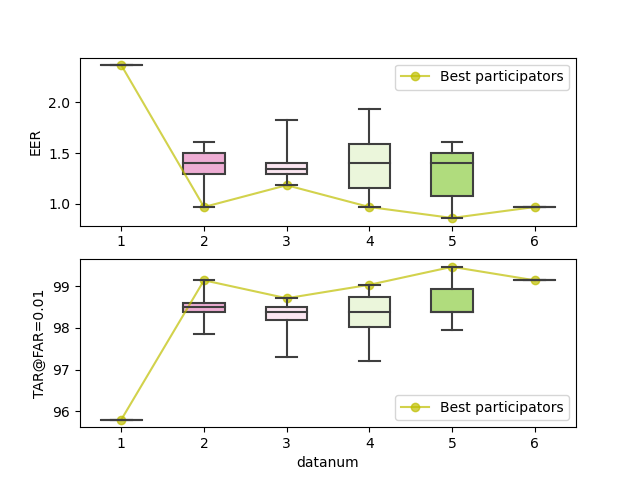}}
\subfigure[]{\label{fig_5c}     
\includegraphics[width=0.25\textwidth]{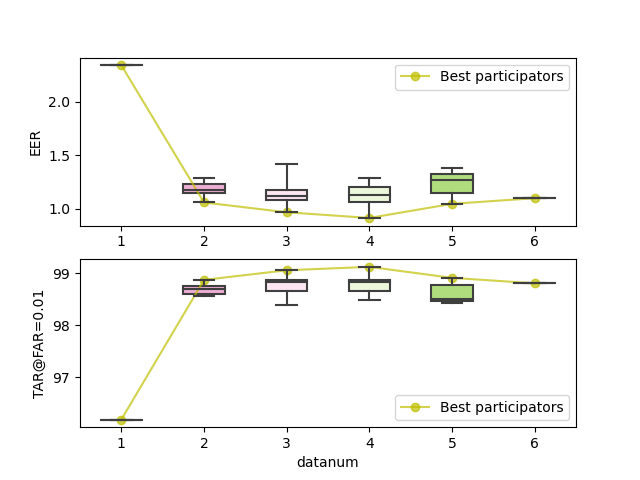}}
\subfigure[]{\label{fig_5d}     
\includegraphics[width=0.25\textwidth]{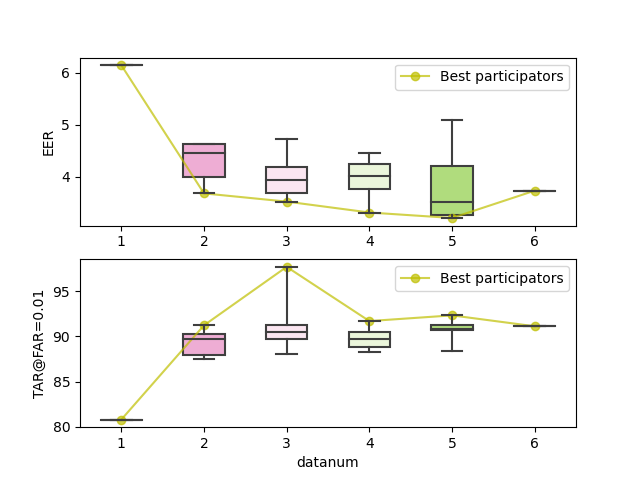}}
\subfigure[]{\label{fig_5e}     
\includegraphics[width=0.25\textwidth]{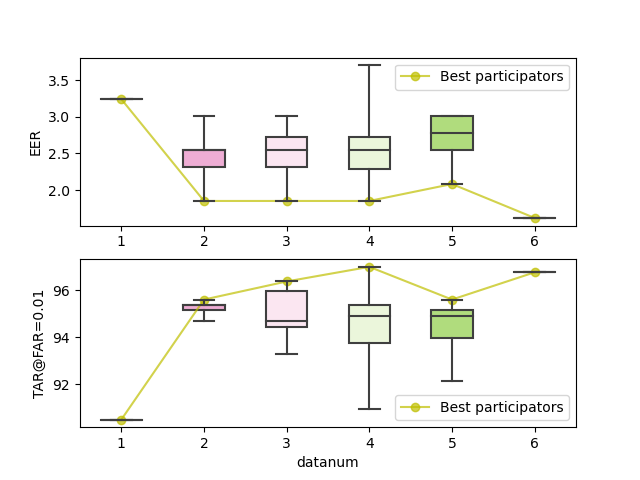}}
\subfigure[]{\label{fig_5f}     
\includegraphics[width=0.25\textwidth]{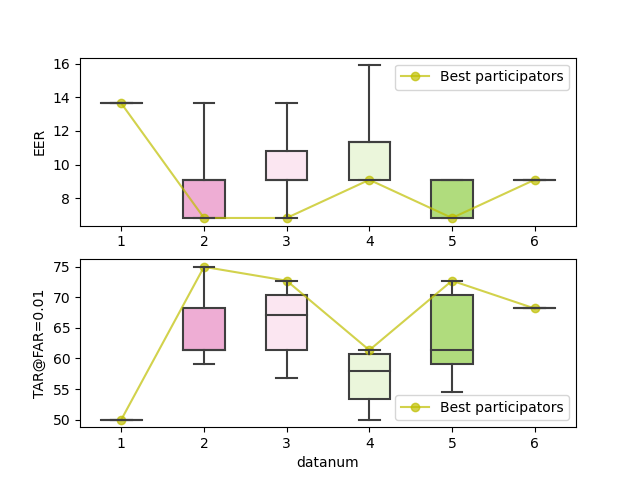}}
\caption{Boxplots of different dataset combinations. where (a) is the EER and TAR(@FAR=0.01) results for different datasets combinations on FV-USM (b) is on HKPU (c) is on NUPT-FV (d) is on SDUMLA (e) is on UTFVP and (f) is on VERA.}
\label{fig_5}
\end{figure*}

\subsection{Experimental Setup}
\subsubsection{Dataset} This work allocates FV-USM, HKPU-FV, NUPT-FV, SDUMLA-HMT, UTFVP, and VERA datasets among six clients. The setup simulates a real-life FL scenario across multiple institutions. Additionally, finger vein recognition in this work is framed as an open-set task, implying that test set have different users from training set. The division between training and test sets follows an 8:2 split ratio. For instance, the FV-USM dataset encompasses 492 classes, of which the first 394 classes constitute the training set, while the remaining classes serve as the test set. Lastly, a limited number of images from the THU-FVFDT \cite{c:38} dataset are selected to create an independent dataset within server.

\subsubsection{Evaluation Metrics} Following the formal finger vein recognition works, we evaluate the Equal Error Rate (EER) and TAR@FAR=0.01 which means the True Acceptance Rate (TAR) at False Acceptance Rate (FAR) equals to 0.01. And the EER means the TAR while TAR equals FRR.

\subsubsection{Experimental Settings} All experiments described in this paper were conducted on an eight-core Windows 11 device featuring an Intel i7-9700K processor. The codebase was implemented base on PyTorch framework, and model training utilized RTX 2080Ti GPUs. The local and federated channels of model in the experiments were set to ResNet50 and ResNet18 \cite{c:36}, respectively. On clients, the epoch of local training is set up 3. And the round of federated learning is set up 10. The base learning rate is 0.01 and the batchsize size is 16.

\subsection{Result and Analysis}
This section presents a series of four experiments conducted to assess the efficacy of PAFedFV. The experiments were formulated to achieve the following objectives. (1) Investigate the impact of non-IID finger vein data on PAFedFV using varying dataset combinations. (2) Validate the effectiveness of PAFedFV on individual clients. (3) Examine the performance of each module within PAFedFV through ablation experiments. (4) Compare PAFedFV against existing methods. Experiment (1) entails the utilization of diverse dataset combinations, while experiments (2) through (4) select the combination of six datasets. The experimental results are detailed and analyzed subsequently.

\subsubsection{Heterogeneity influence} Figure \ref{fig_5} demonstrates the performance of PAFedFV with different dataset combinations. The datanum denotes the number of participating clients. For example, in Figure \ref{fig_5a}, it is considered as SOLO training on FV-USM when datanum = 1. When datanum=2, the box shows five results generated by pairing FV-USM with other data sets. When datanum=3, 4, and 5, the results of 10, 10, and 5 combinations are shown, respectively. And when datanum=6, all datasets participate in federated learning.

By analyzing Figure \ref{fig_5}, the following conclusions can be drawn. 1) Heterogeneity among different clients' data significantly impacts the performance of federated learning. Thus, having more participants does not necessarily lead to better performances for the federated client. Optimal results can be achieved by selecting the right combinations of participants. For instance, when all clients participated (datanum=6), the best performance was not observed across all datasets except UTFVP. 2) Different combinations of participants, even when having the same datanum, can result in substantial variations in outcomes. Particularly, two different participant combinations, both with datanum = 3, respectively yielded the best and worst performances on the FV-USM dataset. 3) The performance of each client does not experience an indefinite improvement with increasing participants. As illustrated in Figure \ref{fig_5a} and \ref{fig_5c}, performance ceases to improve further after reaching the optimal outcome, even with continued increases in participants. Moreover, the performance even exhibits a decline when datanum = 6. To provide a more comprehensive analysis and demonstration, detailed results for datanum = 2 are recorded in Table \ref{table1}. Based on our analysis of the experimental results, we have identified two patterns that can provide valuable insights for future researches. 

\begin{table}[t]
\centering
\resizebox{.85\columnwidth}{!}{
\begin{tabular}{lrrrrrr}
\toprule
\multirow{2}{*}{Client} & \multicolumn{6}{c}{Cooperator}                                                 \\ \cline{2-7} 
                        & USM  & HKPU          & NUPT       & SDUMLA        & UTFVP         & VERA       \\ \midrule
USM                     & -    & {\underline {0.68}}    & 0.61       & \textbf{0.54} & 0.61          & {\underline {0.68}} \\
HKPU                    & 1.29 & -             & 1.40       & 1.51          & \textbf{0.97} & {\underline {1.61}} \\
NUPT                    & 1.18 & 1.23          & -          & \textbf{1.06} & 1.15          & {\underline {1.28}} \\
SDUMLA                  & 3.99 & \textbf{3.68} & {\underline {4.62}} & -             & 4.46          & {\underline {4.62}} \\
UTFVP                   & 2.32 & \textbf{1.85} & 2.55       & {\underline {3.01}}    & -             & 2.55       \\
VERA                    & 9.09 & {\underline {13.64}}   & 9.09       & \textbf{6.82} & 9.09          & -          \\ \midrule
\end{tabular}}
\caption{The EERs when datanum = 2.}
\label{table1}
\end{table}

\begin{itemize}
\item Collaborations involving clients with smaller datasets tend to yield inferior performance. For example, VERA, the smallest dataset in this work, is the worst collaborator when it collaborates with USM, HKPU, NUPT, and SDUMLA. Notably, VERA did not emerge as the optimal collaborator for any dataset.
\item The distributed differences among finger vein datasets, such as variations in finger position and image light intensity, can exert an influence on federated learning. For instance, NUPT-FV, the largest dataset in this work, exhibited subpar performance when collaborating with other clients due to its differing image collection direction. On the other hand, HKPU, SDUMLA and UTFVP, with similar finger positions, demonstrated strong compatibility occurring the best combination several times.
\end{itemize}

\begin{table}[t]
\centering
\resizebox{0.85\columnwidth}{!}{
\begin{tabular}{llrrrrrrrrr}
\toprule
\multirow{2}{*}{\textbf{Metrics}}                                        & \multirow{2}{*}{\textbf{Methods}} & \multicolumn{6}{c}{\textbf{Datasets}}      \\ \cline{3-8} 
                     &             & USM   & HKPU  & NUPT  & SDUMLA & UTFVP & VERA  \\ \midrule
\multirow{3}{*}{EER} & SOLO        & 1.09 & 2.37  & 2.34  & 6.14   & 3.24  & 13.64 \\
                     & PAFedFV     & 0.68  & 0.97  & 1.10  & 3.73    & 1.62  & 9.09  \\ \cline{2-8} 
                     & Centralized & \multicolumn{6}{c}{2.05}                       \\ \midrule
\multirow{3}{*}{\begin{tabular}[c]{@{}c@{}}TAR@FAR\\ =0.01\end{tabular}} & SOLO                              & 98.64 & 95.81 & 96.19 & 80.79 & 90.51 & 50.00 \\
                     & PAFedFV     & 99.46 & 99.14 & 98.81 & 91.13  & 96.76 & 68.18    \\ \cline{2-8} 
                     & Centralized & \multicolumn{6}{c}{95.54}                      \\ \midrule
\end{tabular}}
\caption{The results of comparisons when datanum = 6.}
\label{table2}
\end{table}

\begin{table*}[t]
\centering
\resizebox{0.6\textwidth}{!}{
\begin{tabular}{lccc|rrrrrr}
\toprule
\multirow{2}{*}{\textbf{Metrics}} &
  \multicolumn{3}{c|}{\textbf{Module}} &
  \multicolumn{6}{c}{\textbf{Datasets}} \\ \cline{2-10} 
 &
  \begin{tabular}[c]{@{}c@{}}Asynchronous\\ Training\end{tabular} &
  $L_{total}$ &
  \begin{tabular}[c]{@{}c@{}}Personal\\ Aggregation\end{tabular} &
  USM &
  HKPU &
  NUPT &
  SDUMLA &
  UTFVP &
  VERA \\ \midrule
\multirow{5}{*}{EER} &
  × &
  × &
  × &
  3.13 &
  4.09 &
  2.25 &
  12.28 &
  10.65 &
  22.73 \\
 &
  $\surd$ &
  × &
  × &
  2.38 &
  3.55 &
  1.69 &
  5.77 &
  7.41 &
  \textbf{9.09} \\
 &
  $\surd$ &
  × &
  $\surd$ &
  1.57 &
  3.87 &
  1.16 &
  4.72 &
  4.18 &
  \textbf{9.09} \\
 &
  $\surd$ &
  $\surd$ &
  × &
  \textbf{0.68} &
  1.18 &
  1.18 &
  4.20 &
  2.31 &
  11.36 \\
 &
  $\surd$ &
  $\surd$ &
  $\surd$ &
  \textbf{0.68} &
  \textbf{0.97} &
  \textbf{1.10} &
  \textbf{3.73} &
  \textbf{1.62} &
  \textbf{9.09} \\ \midrule
\multirow{5}{*}{TAR@FAR =0.01} &
  × &
  × &
  × &
  92.45 &
  91.61 &
  96.48 &
  61.63 &
  62.03 &
  36.36 \\
 &
  $\surd$ &
  × &
  × &
  94.97 &
  92.58 &
  97.84 &
  86.35 &
  81.48 &
  \textbf{68.18} \\
 &
  $\surd$ &
  × &
  $\surd$ &
  97.82 &
  92.58 &
  98.70 &
  87.61 &
  84.49 &
  59.09 \\
 &
  $\surd$ &
  $\surd$ &
  × &
  99.39 &
  98.82 &
  98.68 &
  88.29 &
  \textbf{96.99} &
  61.36 \\
 &
  $\surd$ &
  $\surd$ &
  $\surd$ &
  \textbf{99.46} &
  \textbf{99.14} &
  \textbf{98.81} &
  \textbf{91.13} &
  96.76 &
  \textbf{68.18} \\ \midrule
\end{tabular}}
\caption{The result of ablation experiment.}
\label{table3}
\end{table*}

\begin{figure*}[t]
\centering
\subfigure[]{\label{fig_6a}
\includegraphics[width=0.28\textwidth]{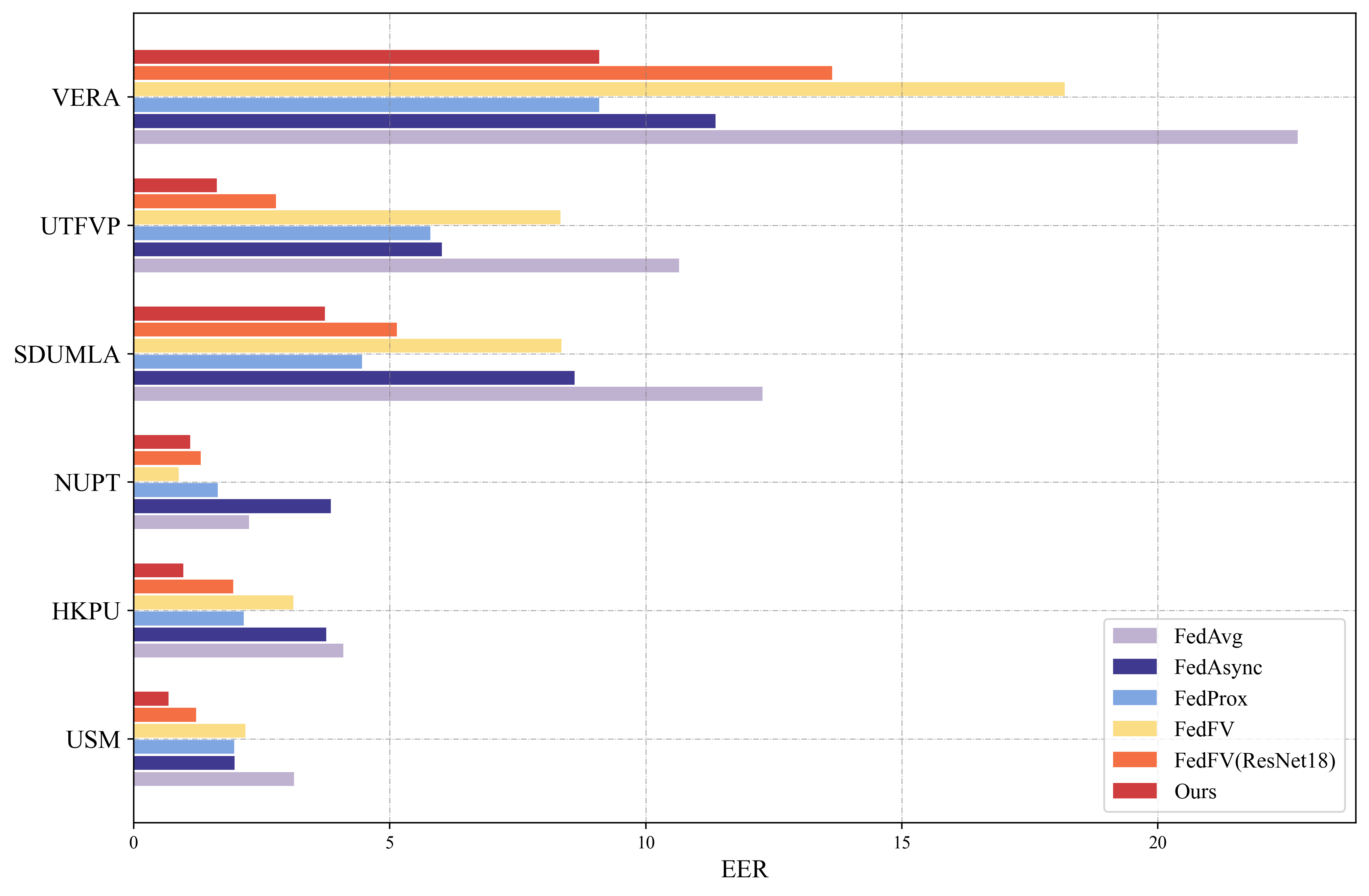}}    
\subfigure[]{\label{fig_6b}     
\includegraphics[width=0.28\textwidth]{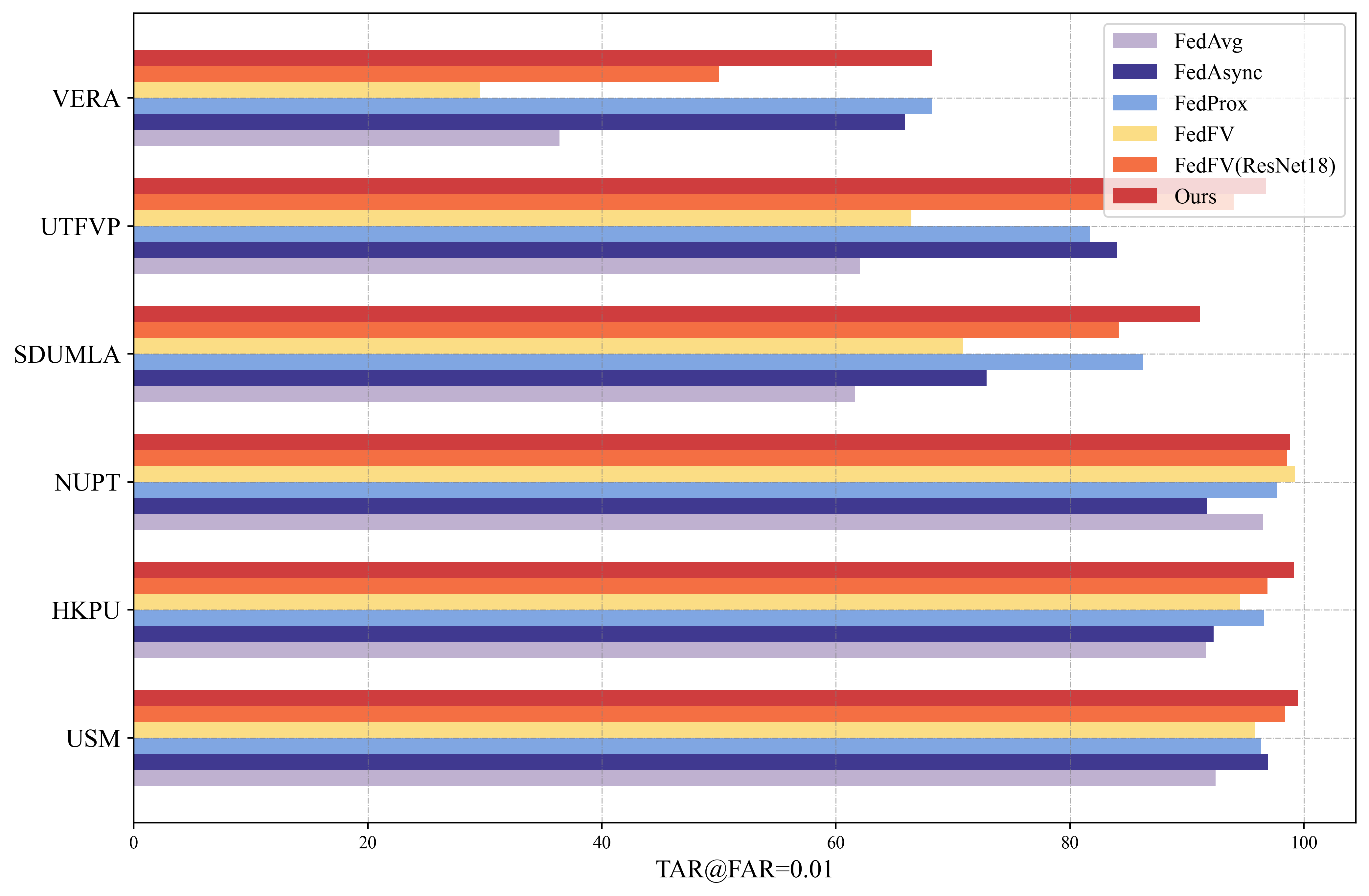}}
\caption{The results of comparation experiment. where (a) is the EER results (b) is the TAR@FAR=0.01.}
\label{fig_6}
\end{figure*}

\begin{figure*}[htbp!]
\centering
\subfigure[]{\label{fig_7a}
\includegraphics[width=0.15\textwidth]{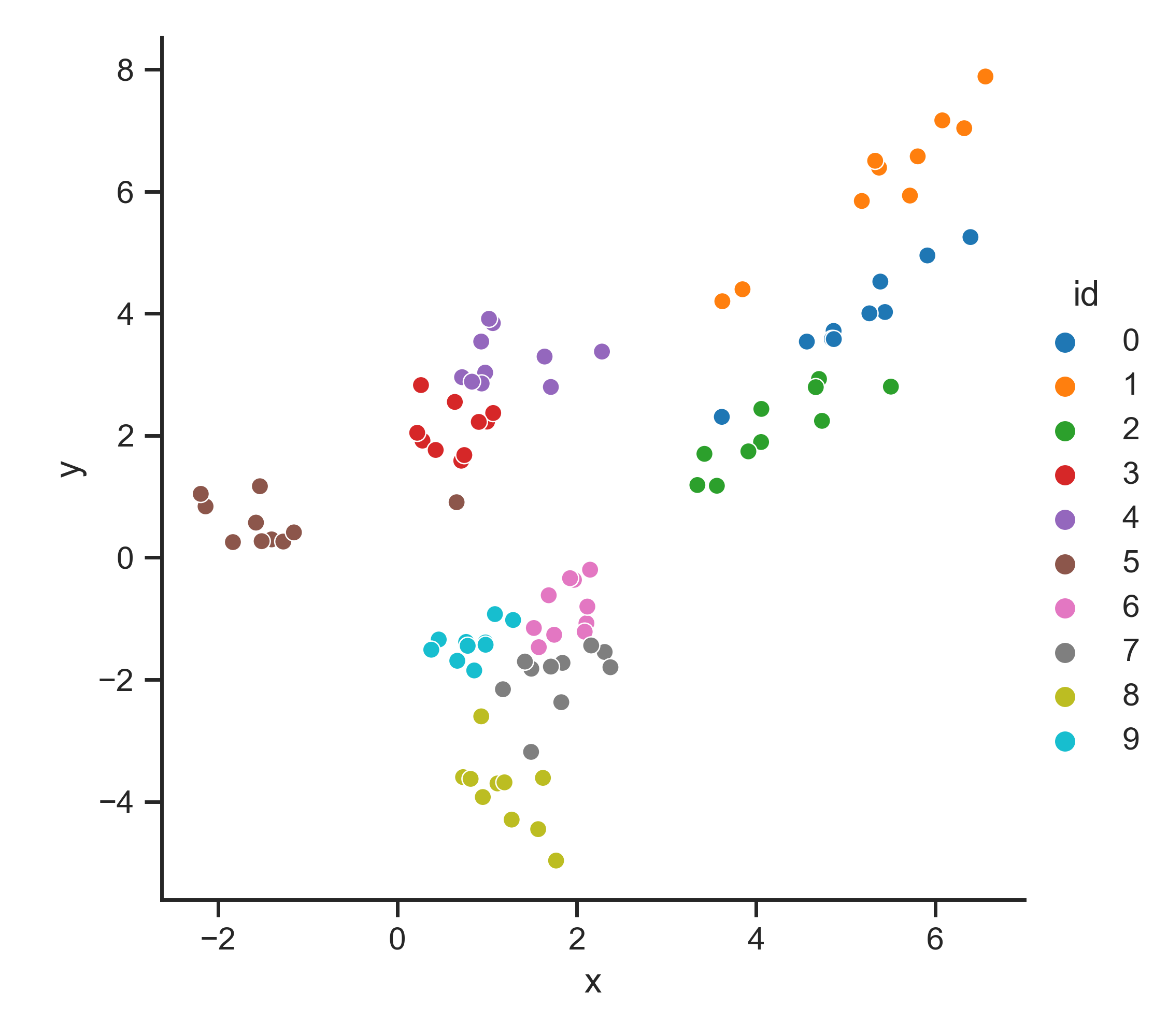}}    
\subfigure[]{\label{fig_7b}     
\includegraphics[width=0.15\textwidth]{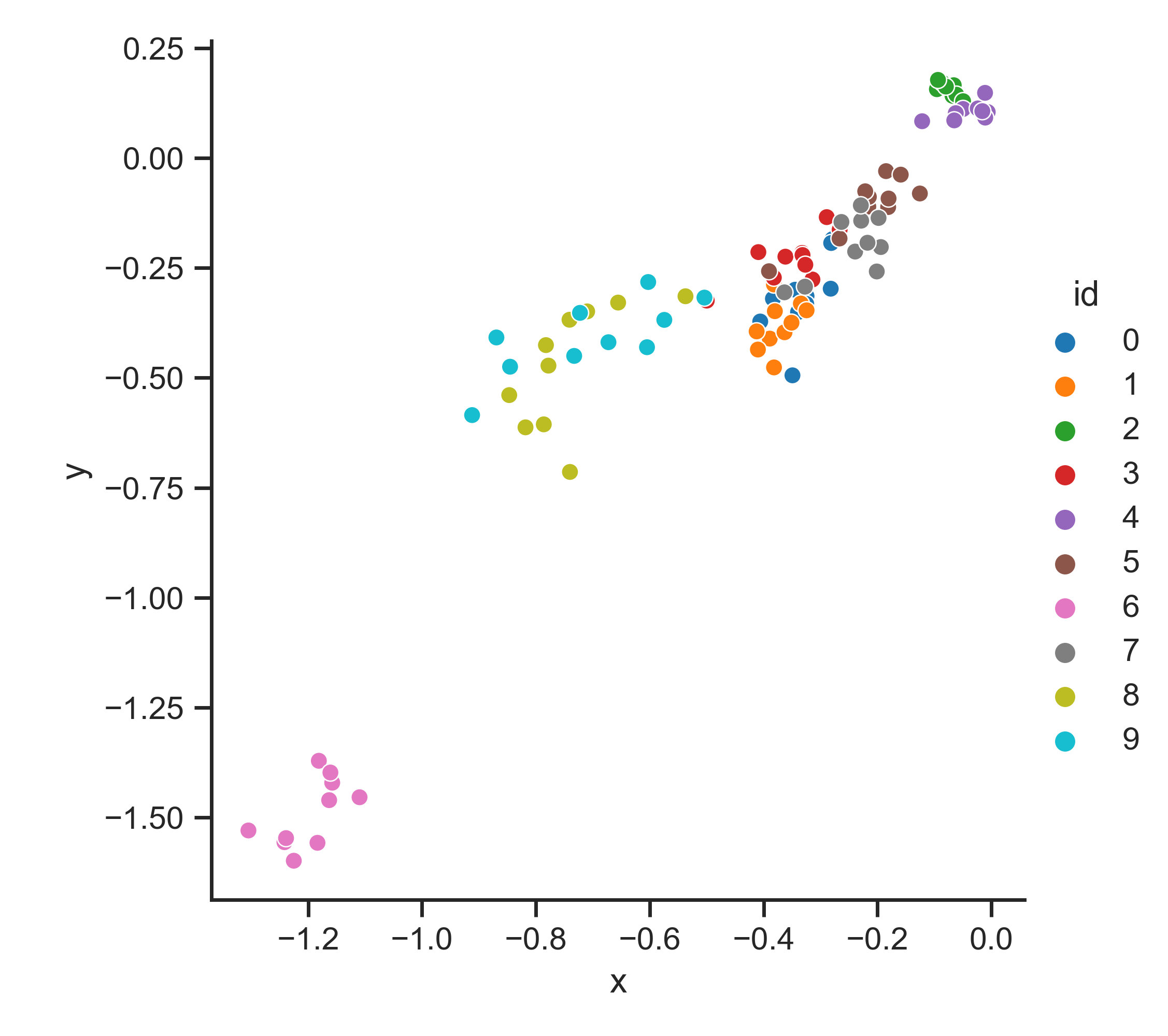}}
\subfigure[]{\label{fig_7c}     
\includegraphics[width=0.15\textwidth]{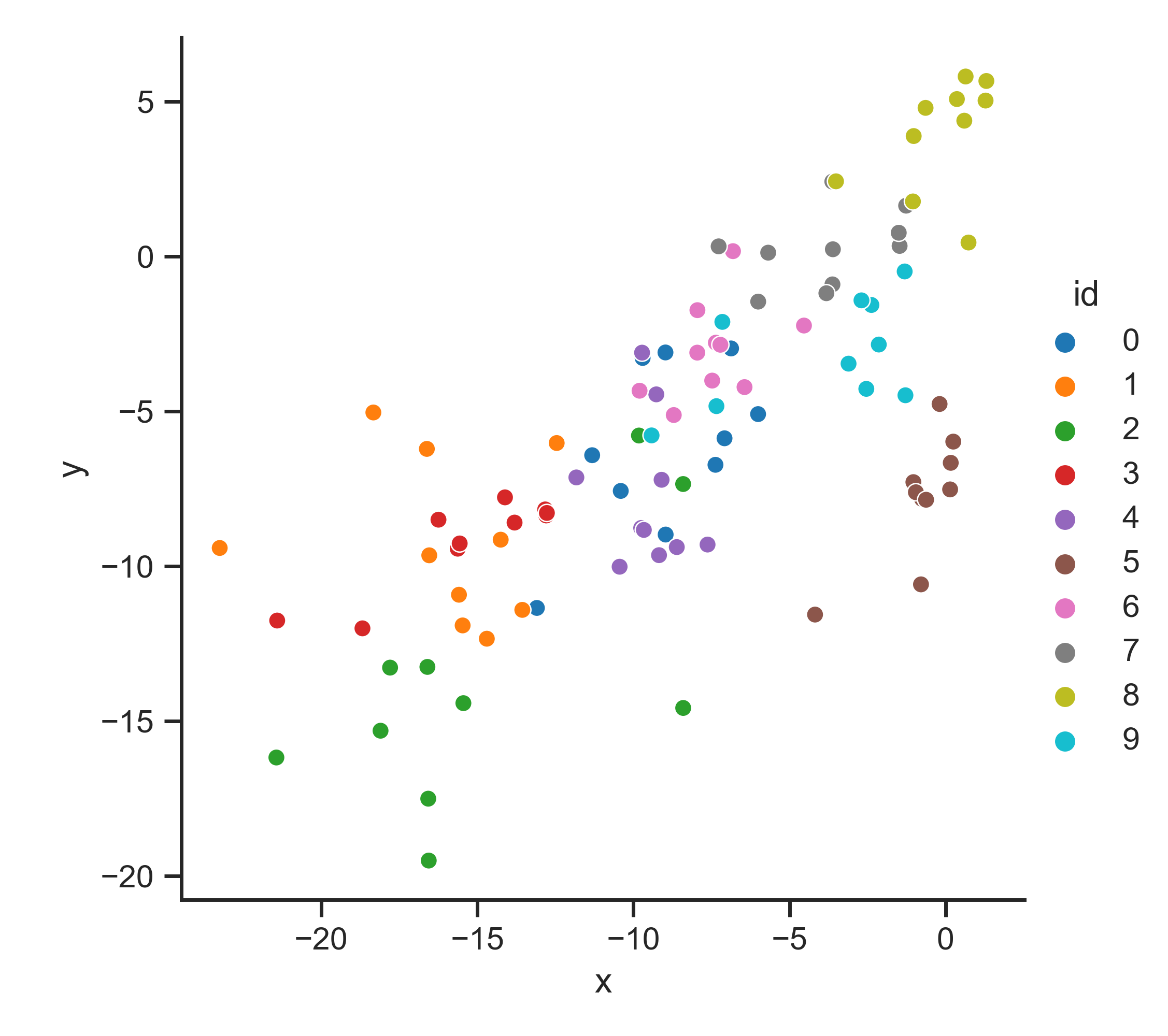}}
\subfigure[]{\label{fig_7d}     
\includegraphics[width=0.15\textwidth]{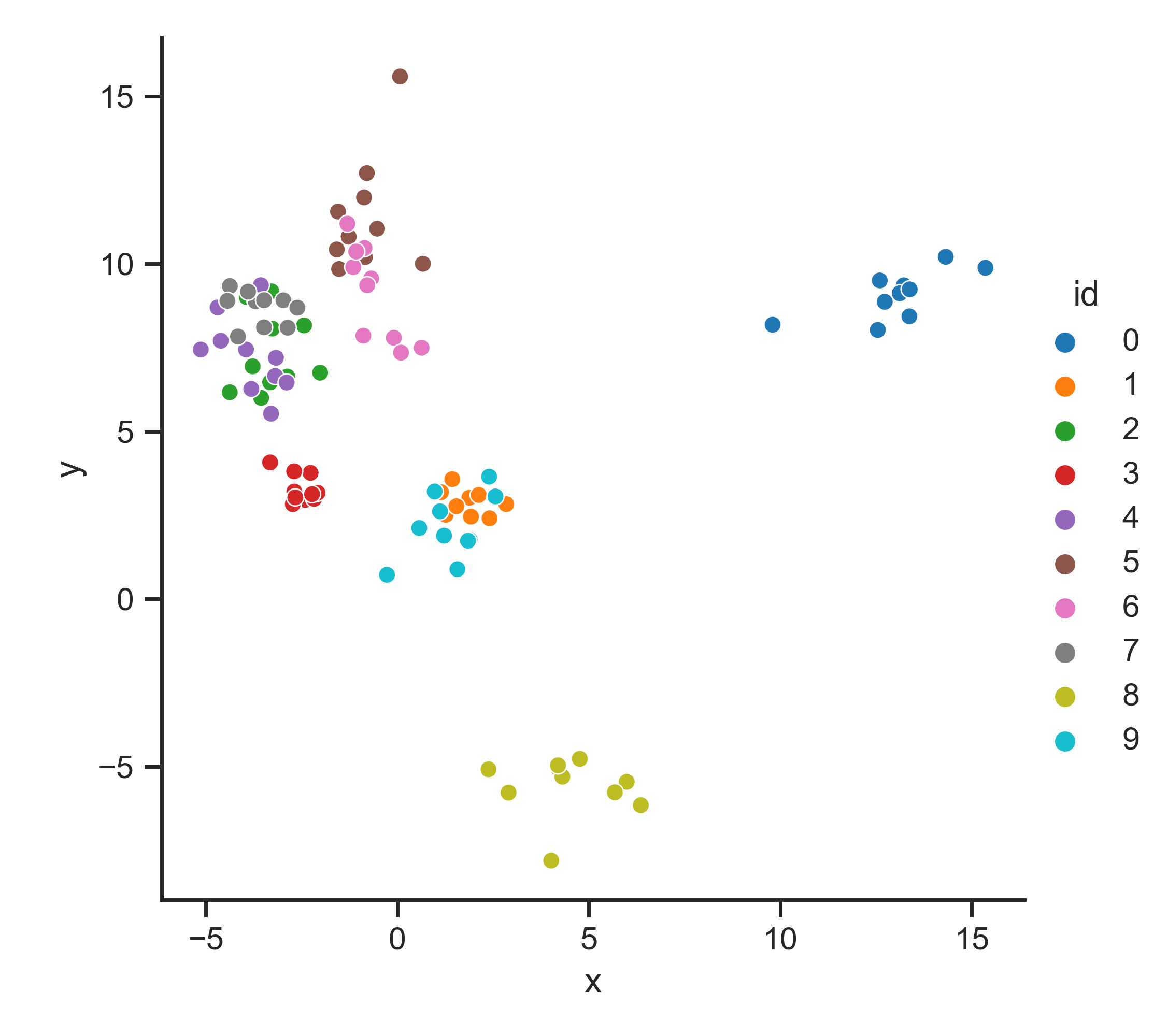}}
\subfigure[]{\label{fig_7e}     
\includegraphics[width=0.15\textwidth]{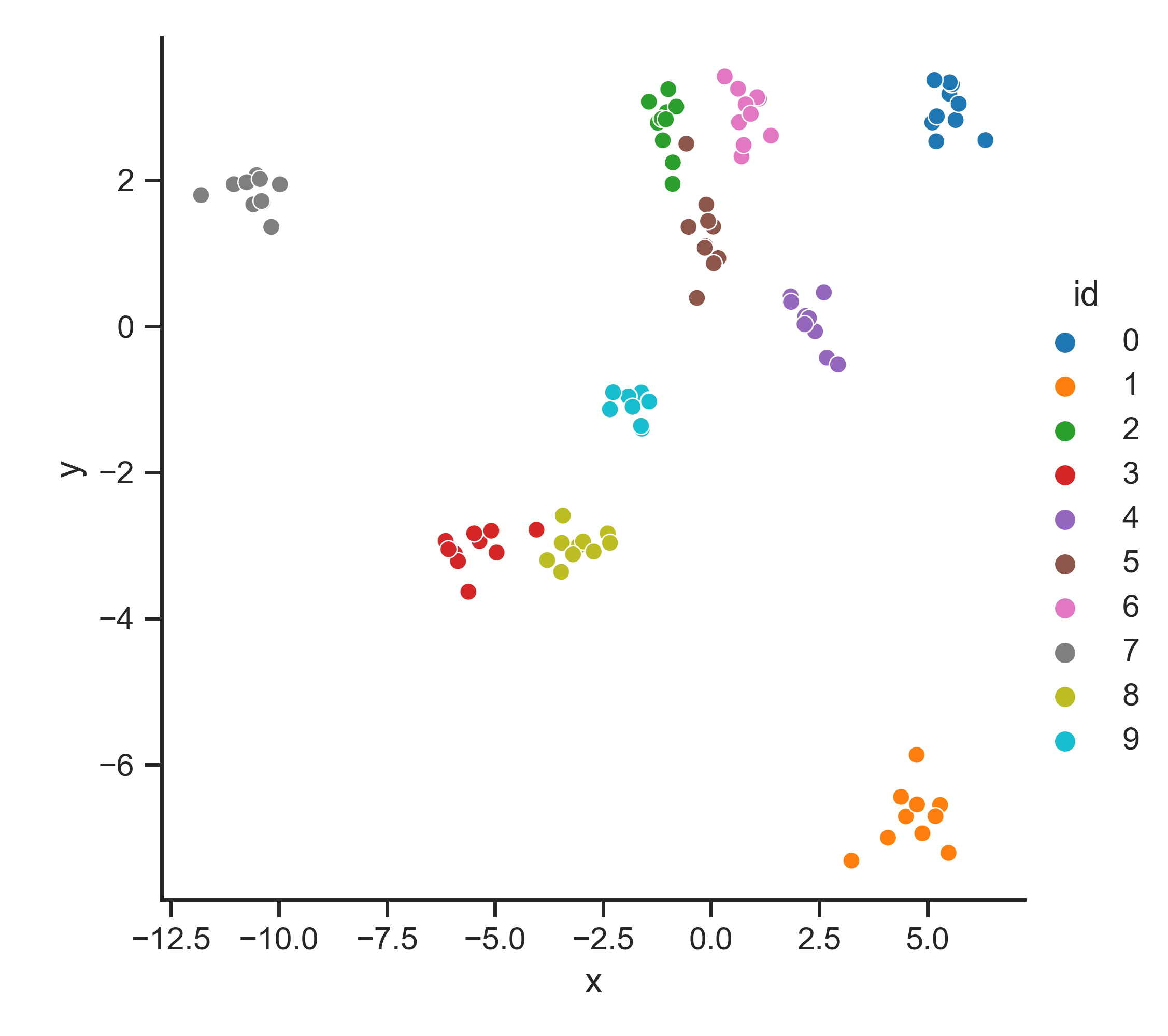}}
\subfigure[]{\label{fig_7f}     
\includegraphics[width=0.15\textwidth]{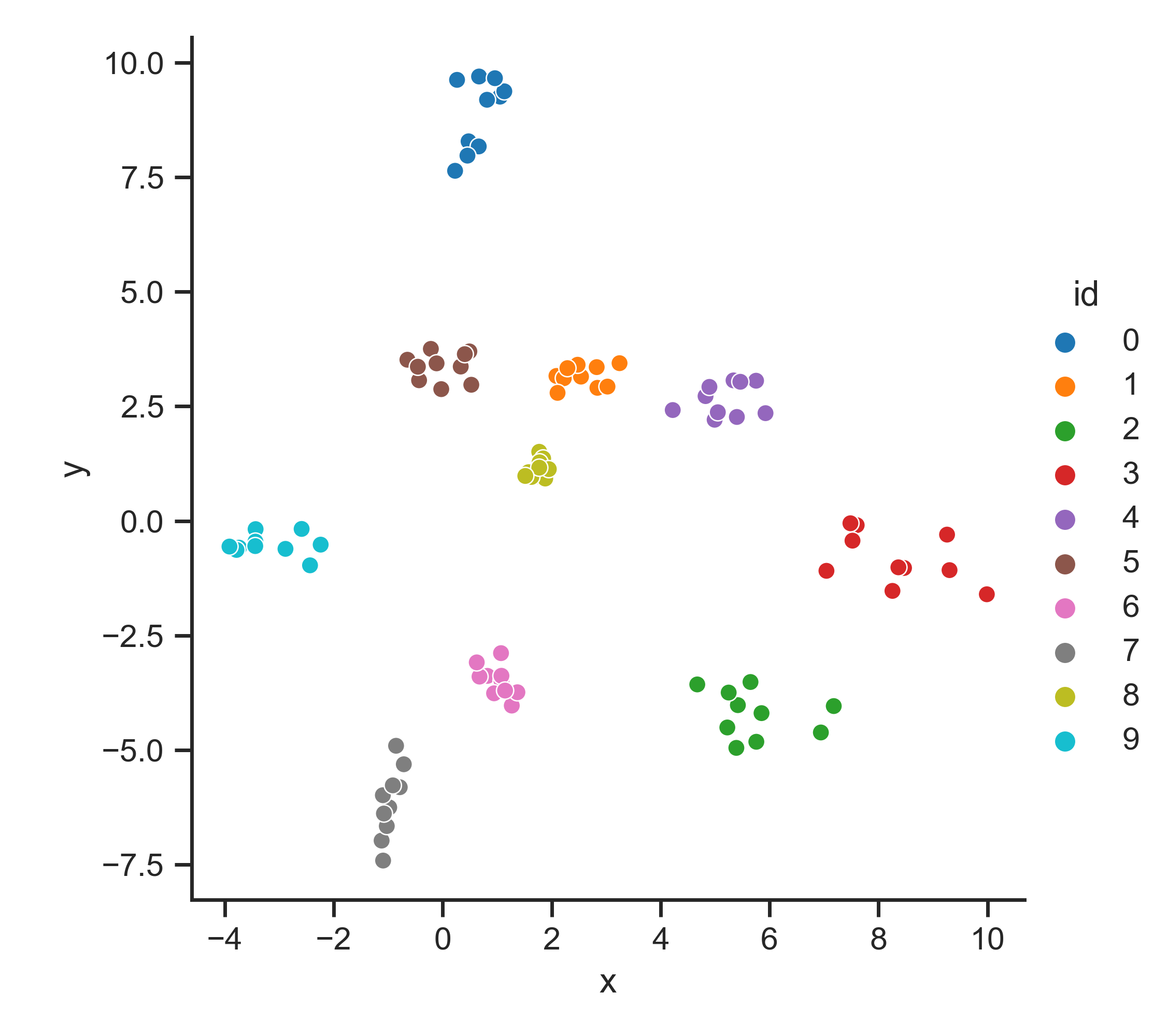}}
\caption{The figure shows the finger vein representation distributions of 10 classes in NUPT-FV dataset. The models are trained by (a) FedAvg, (b) FedAsync (c) FedProx (d) FedFV (MobileNet), (e) FedFV (ResNet) and (f) PAFedFV.}
\label{fig_7}
\end{figure*}

\subsubsection{Effectiveness of PAFedFV} To assess the efficacy of PAFedFV, it was compared with SOLO and Centralized training. The outcomes are presented in Table \ref{table2}, showcasing PAFedFV's superior performance. Among them, its efficacy is most pronounced on VERA, where it reduces the EER by 4.55 and enhances TAR(@FAR=0.01)  by 18.18. These improvements suggest that PAFedFV provides more substantial enhancements for clients with smaller datasets. Moreover, PAFedFV even surpasses Centralized training on USM, HKPU, NUPT, and UTFVP datasets. Specifically, on USM, PAFedFV yields the most exceptional results with reducing EER by 1.37 and enhancing TAR(@FAR=0.01) by 3.92 compared to Centralized training. This observation, corroborated by the findings in Figure \ref{fig_5}, underscores the detrimental impact of distributed differences among non-IID finger vein datasets on Centralized training. In conclusion, PAFedFV exhibits high robustness and accuracy and effectively.

\subsubsection{Ablation Experiment} The modules of Asynchronous Training, $L_{total}$, and Personal Aggregation are integrated sequentially into the foundational FedAvg framework for comparative analysis. The results are presented in Table \ref{table3}. The Asynchronous Training module signifies the incorporation of asynchronously trained local channels into the personalized model. Upon its addition, a substantial enhancement is observed in EER and TAR(@FAR=0.01) across the finger vein datasets. Notably, on VERA, EER decreases by 13.64 and TAR(@FAR=0.01) increases by 31.82. The $L_{total}$ is designed in this work to optimize finger vein recognition model, instead of the traditional cross-entropy loss function. The experiment results show that the performance of almost all clients is further improved with $L_{total}$. Particularly, on UTFVP, the most significant advancement is seen, reducing EER by 5.1 and elevating TAR(@FAR=0.01) by 15.51. The Personal Aggregation module, designed to replace the traditional model averaging approach, leads to a notable performance boost for nearly all clients, with an average EER reduction of 0.62 and an average TAR(@FAR=0.01) improvement of 1.66. The experiment results show that these modules effectively improve the performance of the finger vein recognition model and solve the heterogeneity among finger vein datasets.

\subsubsection{Comparison with other methods} In conclusion, this paper is compared with four advanced methods, FedAvg \cite{c:6}, FedAsync \cite{a:44}, FedProx \cite{r:45}, FedFV \cite{r:7}. The FedAvg is the most classical federated learning method. The FedAsync explores the asynchronized federated learning which is similar to this paper. And the FedProx is the very standard personalized federated learning framework. Additionally, two FedFV variants are implemented base respectively on MobileNet ~\cite{c:37} and ResNet models for a comprehensive comparative analysis.

As demonstrated in Figure \ref{fig_6}, PAFedFV exhibits superior performance compared to five existing methods on all datasets, which yields 3.98, 1.47, 6.32, 3.07, and 1.32 increase on average in EER and 16.19, 5.26, 18.82, 8.3, and 4.46 decrease on average in TAR(@FAR=0.01), respectively.  And to further compare the performance of the four methods, their results on the NUPT-FV are visualized in Figure \ref{fig_7}. As is shown in Figure \ref{fig_7d}, the PAFedFV also achieves the best results. The finger vein representations in PAFedFV are more capable of distinguishing different identities than other methods. This corroborates the superiority of PAFedFV in accuracy and robustness.

\section{Conclusion}
In this work, we propose a Personalized and Asynchronous Federated Learning for Finger Vein Recognition framework, PAFedFV. This framework enhances the applicability of federated learning for finger vein recognition by designing personalized model aggregation to accommodate to heterogeneous non-IID finger vein data. Furthermore, an asynchronized training module is implemented within PAFedFV to effectively utilize clients' waiting periods. Moreover, this paper reports extensive experiments conducted on six datasets. This paper also reports the performance of PAFedFV on six datasets. The experiment results show that our method is of highly accuracy and robustness, outperforms existing methods. Additionally, we investigate the impact of non-IID finger vein data with federated learning, and summarize the regularities that can be referenced by future research.

\subsubsection{Acknowledgements} This work was supported in part by the National Natural Science Foundation of China under Grant 62272242, and the Postgraduate Research \& Practice Innovation Program of Jiangsu Province under Grant KYCX23\_1074 and Grant SJCX23\_0268.


\bibliographystyle{named}
%\bibliography{ijcai24}

\end{document}